\documentclass{article}

\usepackage[final]{neurips_2019}

\usepackage[utf8]{inputenc} 
\usepackage[T1]{fontenc}    
\usepackage{hyperref}       
\usepackage{url}            
\usepackage{booktabs}       
\usepackage{amsfonts}       
\usepackage{nicefrac}       
\usepackage{microtype}      
\usepackage{multirow}
\usepackage{graphicx}
\usepackage{color, colortbl}

\definecolor{LightGray}{gray}{0.8}

\title{Social and Emotional Skills Training with Embodied Moxie}

\author{%
  Nikki Hurst, OTD, OTR/L\\
  Embodied, Inc.\\
  Pasadena, CA \\
  \texttt{nikki.hurst@embodied.com} \\
  \And
  Caitlyn Clabaugh, Ph.D.\thanks{Corresponding Author.} \\
  Embodied, Inc.\\
  Pasadena, CA \\
  \texttt{caitlyn.clabaugh@embodied.com} \\
  \And
  Rachel Baynes\\
  Embodied, Inc.\\
  Pasadena, CA \\
  \texttt{rachel.baynes@embodied.com} \\
  \And
  Jeff Cohn, Ph.D.\\
  University of Pittsburgh\\
  Pittsburgh, PA  \\
  \texttt{jeffcohn@pitt.edu} \\
  \And
  Donna Mitroff, Ph.D.\\
  The Kidvocate Group, LLC\\
  Oakland, CA\\
  \texttt{donnamitroff@earthlink.net} \\
  \And
  Dr. Stefan Scherer\\
  Embodied, Inc.\\
  Pasadena, CA \\
  \texttt{stefan@embodied.com}
}

\begin{document}

\maketitle

\begin{abstract}
  We present a therapeutic framework, namely \textit{STAR Framework}, that leverages established and evidence-based therapeutic strategies delivered by the Embodied Moxie, an animate companion to support children with mental behavioral developmental disorders (MBDDs). This therapeutic framework jointly with Moxie aims to provide an engaging, safe, and secure environment for children aged five to ten years old. Moxie delivers content informed by therapeutic strategies including but not limited to naturalistic Applied Behavior Analysis, graded cueing, and Cognitive Behavior Therapy. Leveraging multimodal input from a camera and microphones, Moxie is uniquely positioned to be a first-hand witness of a child's progress and struggles alike. Moxie measures skills captured in state-of-the-art assessment scales, such as the Social Responsiveness Scale and Social Skill Improvement Scale, and augments those measures with quantitatively measured behavior skills, such as eye contact and language skills. While preliminary, the present study (N=12) also provides evidence that a six-week intervention using the STAR Framework and Moxie had significant impact on the children's abilities. We present our research in detail and provide an overview of the STAR Framework and all related components, such as Moxie and the companion app for parents.
\end{abstract}

\section{Introduction}

Mental, Behavioral, and Developmental Disorders (MBDDs) among children continue to impact the lives of many, both children and their loved ones alike. The Centers for Disease Control (CDC) estimates that about one in six children between the ages of 3-17 suffer from one or more MBDDs in any given year\footnote{\url{https://www.cdc.gov/childrensmentalhealth/features/kf-childrens-mental-health-report.html}}. Further, the CDC estimates that more than 230 billion USD are spent each year on MBDDs, both by insurances as well as by families out of pocket \citep{wang2013healthcare}. MBDDs are characterized as serious changes in the ways children learn, behave, or handle their emotions. While symptoms can manifest in early childhood, typically children are not diagnosed until they reach school age. Once diagnosed, MBDDs can be managed and treated through intensive evidence-based care \citep{case2008evidence}. However, many children with MBDDs may never be diagnosed or recognized as being affected.

The current methods used to assess social-emotional functioning, such as self-report surveys, semi-structured interviews, and clinician judgment, suffer from several shortcomings that can critically impact and limit the diagnosis, and therefore, treatment of MBDDs. First, a lot of assessments rely heavily, if not entirely, on patient and parent or guardian feedback; such self-reporting is not only subjective, it also requires extensive involvement and copious amounts of time from the already impacted families. Second, many of these standard methods require considerable training to establish inter-rater reliability \citep{kobak2004rater}, which can be time-consuming to administer and score, and can often require multiple sources of information to make accurate ratings \citep{ho1998two}. Third, there are many sources of bias and variability that influence ratings, such as gender, ethnicity, personality, mood, and other environmental factors, that often cannot be eliminated or accounted for. Fourth, from a psychometric perspective, symptom-rating scales are not ideal because they often employ vague rating systems (e.g., ``mild'', ``moderate'' and ``severe'' categories from \citep{lukoff1986manual}) that may be insensitive to subtle changes in symptoms and their severity over time \citep{eckert1996comparison}. Fifth, in non-research clinical settings, quantitative measures of social-emotional behavior and functioning are rarely accessible. Instead, clinicians are required to use their own subjective judgment as part of clinical interviews to determine if patients have improved or worsened. Thus, the state-of-the-art in clinical research and practice could considerably benefit from easy to access quantitative assessments of psychological health and social or emotional functioning. We aim to provide such additional information to parents and caregivers alike through the employment of the Embodied Moxie and the STAR Framework.

In fact, advances in computational behavior analyses and automatic screening technologies provide an opportunity to profoundly impact the science and treatment of MBDDs. Automatic algorithms to detect impairments in social-emotional behavior and functioning from children's nonverbal and verbal behavior have the potential to increase the quantitative assessments, accessibility, and cost-effectiveness of social-emotional behavior assessment and monitoring. Embodied is uniquely positioned to leverage these new technologies to accurately and quantitatively assess a child's status and progress across many dimensions and over extended periods of time in ecologically valid contexts (i.e., at home in the child's room), during constrained and goal driven interactions, and social play with the Embodied robot Moxie. This allows us to individualize training for each child and assess progress in transfer tasks. 

We consider two domains of human behavior and functioning: (1) communicative or social skills and (2) emotional skills. Both domains are equally important for the development of a child and can be further broken down into a number of constructs or units of analysis. With this work we investigate the two domains using two distinct units of analyses: (1) low-level and direct measures of human nonverbal and verbal behavior called behavior markers, including but not limited to facial expressions, gestures, choice of words, timings, and prosody; and (2) higher-level and abstracted measures named skills, including a child's ability to be empathetic, communicate thoughts, and engage in social interactions. In this document we set the stage for the assessment framework employed at Embodied and describe preliminary results from one of our early investigations.








\subsection{Related Work}

\paragraph{Robots for Autism Spectrum Disorder (ASD).}
Among the core symptoms characterizing ASD are impairments in social interaction and communication \citep{american2013diagnostic}. Research has shown that these deficits can be tackled through early and sustained intervention. These interventions if conducted by humans (i.e., therapists, parents, etc.) are severely limited in reach and are costly \citep{amendah2011economic}. Non-human or robotic interventionists could alleviate this lack of resources significantly and have shown to be efficient replacements \citep{begum2016robots,scassellati2012robots}. In fact, there are thousands of peer-reviewed studies that investigate robotics for the use in ASD. These studies show consistent support that robots are engaging kids with ASD and improve their abilities in manifold ways: Increase in social eye gaze, verbalizations, joint attention, turn taking, initiating play, perspective-taking, sharing, social integration, or confidence. To further exemplify the utility of robots to support children with ASD, we introduce some specific studies.

\citet{scassellati2018improving} for example found that repeated social skills training with a robot together with a human caregiver significantly improved social skills, such as joint attention in a within subject design study. Further, \citet{kim2013social} showed that the presence of a robot increases speech output significantly in children with ASD. These effects appear to also persist when the child is required to apply these skills in the context of human-human interaction \citep{scassellati2018improving, kim2012bridging}. Even simple robot pets were shown to improve engagement and increase reciprocal behavior \citep{stanton2008robotic} and positive physical interactions \citep{andreae2014study} in children.

However, research into the efficacy of robots supporting the development of children with ASD suffers from a number of shortcomings. First, ASD is a highly diverse set of symptoms and cognitive, social, as well as emotional abilities vary greatly \citep{lord2012annual}. In addition, access to children with ASD is highly constrained and often study population sizes are small. Further, academic resources are unfortunately often too small to engage many children and access to hardware (i.e., standardized robots with broad range of capabilities) is one of the biggest barriers to success. The lack of abundant hardware further often prohibits studies into the long-term effects. Within this study we make first strides towards alleviating some of these issues.

\paragraph{The effect of Embodiment.}
Embodied agents can be \textbf{embodied} \textit{virtually} through 3-dimensional computer renderings or \textit{physically}, like the Embodied Moxie. Both virtually and physically embodied agents have human-like behaviour (e.g., facial expressions, a voice, etc.) and affordances (e.g., a face, a body, etc.) while interacting with its environment and the human users within. Physically embodied agents have the unique ability to share the physical space with the user. In this case, embodiment affords the agent to engage in ways a virtual agent cannot (e.g., touch, physical manipulation tasks, or making eye contact). This has major advantages in learning non-academic skills, like social, emotional, and motor skills. Human-like affordances, such as a face with two eyes, speech, and gestures, allow an artificial agent to create rapport and increase engagement \citep{gratch2006virtual} and trigger emotional reactions in the human brain through ``hard-wired'' pathways that mirror emotional perception with one's own experience \citep{wicker2003both}. In a meta-review of 65 studies comparing embodied to non-embodied agents we found that, in 93.8\% (N=61) of the studies the embodied agent was at least as efficient as a non-embodied agent and significantly outperformed the non-embodied agent in 78.5\% (N=51) of the studies \citep{deng2019embodiment}. The evaluation combines metrics of quantitative task performance and subjective user experience. \citet{deng2019embodiment} observed that negative results only occur when the agent's design does not match the social role it is supposed to play. The design of the agent scaffolds interactions and is important in setting expectations about the agent such as the way it interacts, its cognitive capabilities, and even its demeanor. When the expected capabilities and intended role mismatched the agents' evaluation was negative. To further exemplify the effect of embodiment, we introduce some specific example studies. 

A study with children from both the Netherlands and Pakistan showed significantly increased affect related gesture cues (e.g., laughter, smiling, etc.) when playing with an embodied robot called iCat as compared to alone play \citep{shahid2014child}. \citet{lee2006physically} observed significantly increased social presence and positivity of agent as well as interaction in a study comparing a physically embodied and virtual Sony Aibo robot dog. Similarly, a study by \citet{komatsu2010comparison} further showed two-fold increased task engagement when comparing an embodied and a virtual robot called PaPeRo. Overall, robots appear to have a captivating effect to children as \citet{jost2012robot} have shown in their work.

\begin{figure}
  \centering
  \includegraphics[width=8cm]{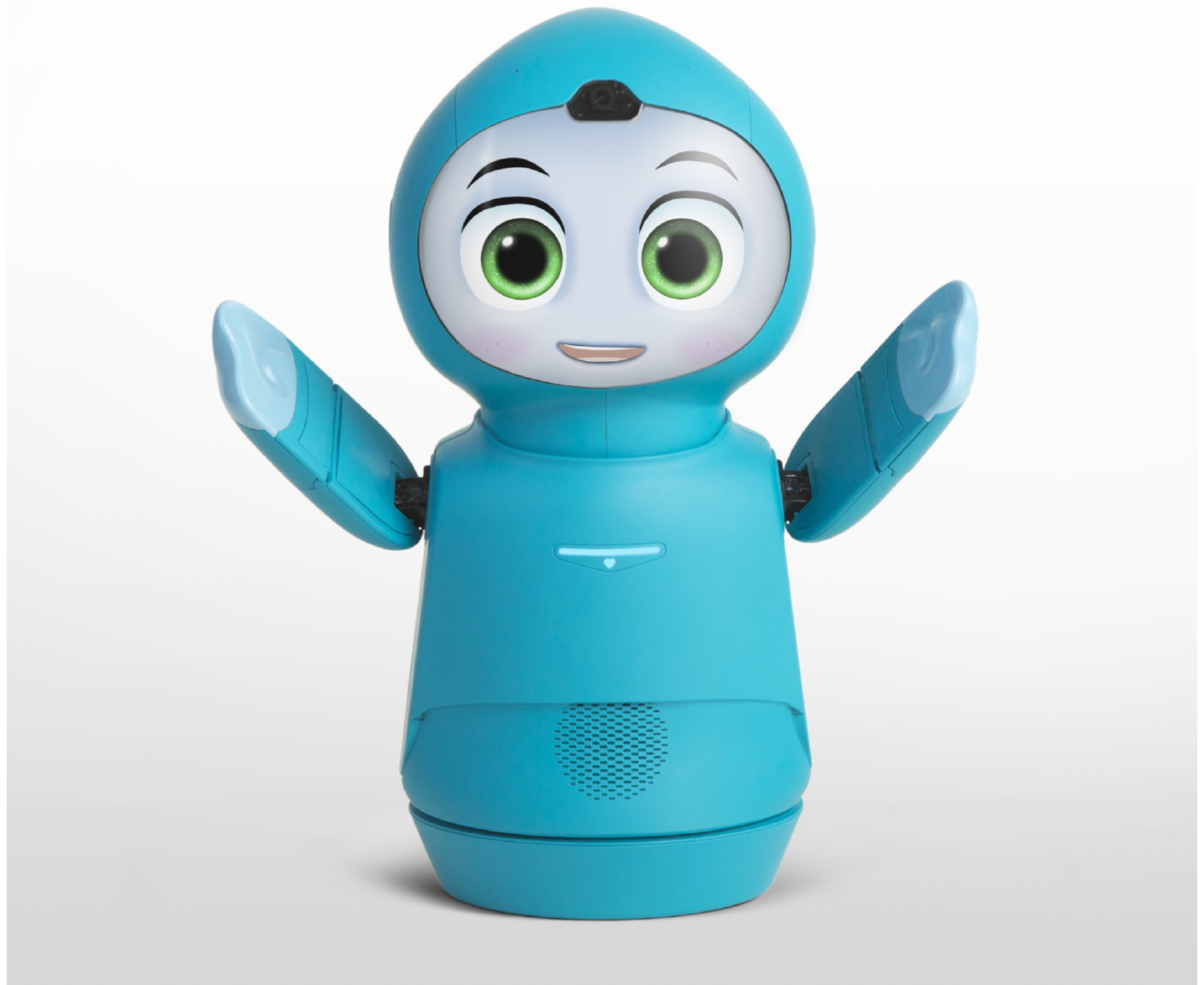}
  \caption{Moxie Robot.}
  \label{fig:moxie}
\end{figure}

\section{Embodied Moxie}
Moxie is a novel companion for children designed to help promote social, emotional and cognitive development through play-based learning. With Moxie, children can engage in meaningful play, every day, with content informed by the best practices in child development and early childhood education.

As an animate companion, Moxie needs to perceive, understand, and react to the physical world that surrounds it and the human users that interact with it. Moxie leverages multimodal information fusion on the edge (i.e., on the robot device itself) to build and track an accurate representation of its surroundings and its users. Using multimodal information, Moxie localizes and prioritizes input from engaged users, recognizes faces, objects, and locations, analyzes facial expressions and voice of the users to assess their affect, mood, and level of engagement, and understands the user's intent, desires, and needs holistically. To enable an animate companion to gain rapport and trust with its human users, it is crucial that the sense-react loop of the system is as tight and fast as possible \citep{gratch2006virtual}. To reduce the time between perceived input and produced output, Moxie leverages a combination of computer vision algorithms, lightweight neural networks, and natural language processing on board. Edge computing allows us to be in control of this sense-react loop and only rely on cloud computing resources when needed. Lastly, we entrust Moxie with supporting and engaging with one of the most precious and vulnerable demographics, our children. Hence, data security is of the utmost importance to us. Edge computing of video data allows us to ensure the required privacy and security for our users as raw video data never leaves the robot. Moxie is visualized in Fig. \ref{fig:moxie}.

\begin{figure}
  \centering
  \includegraphics[width=\textwidth]{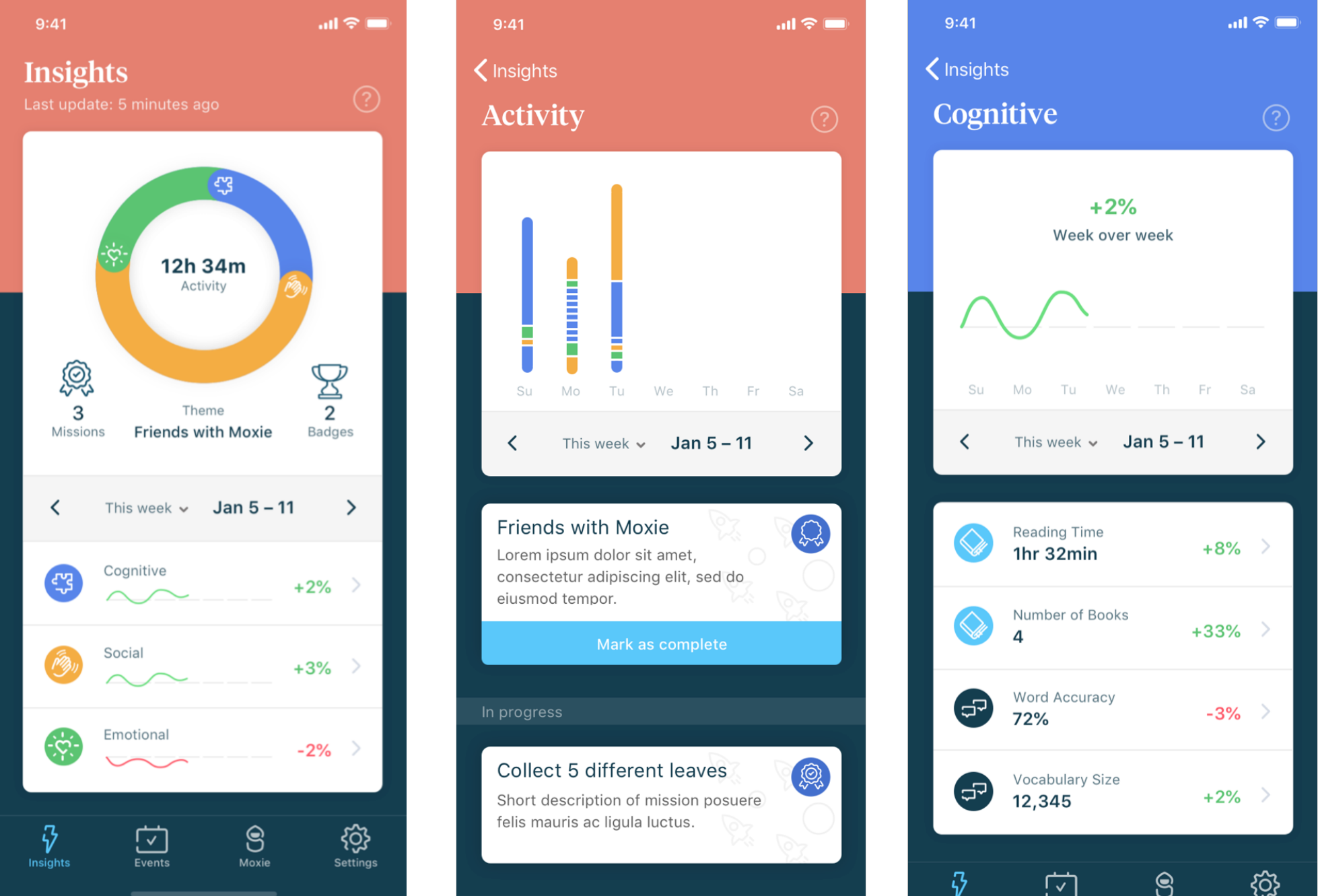}
  \caption{Companion App.}
  \label{fig:app}
\end{figure}

\paragraph{Companion App.} Moxie is accompanied by a companion app that provides parents with quantitative insights of their child's interactions with the robot. Through the app, parents have access to a child's weekly progress, activities, and deep insights into their child's abilities and struggles. In addition, the companion app provides parents with insightful suggestions about their child's development. The companion app is visualized in Fig. \ref{fig:app}. 

\section{STAR Framework}

The STAR Framework, developed by Embodied, is a therapeutic framework designed to allow the robot to help improve social and emotional skills for children with varying challenges and needs. It incorporates tenets of naturalistic Applied Behavior Analysis (nABA) therapy, graded cueing, and cognitive behavioral therapy (CBT) in order to help children make progress towards specific goals and generalize learned skills to their social interactions in the real world. The STAR Framework also allows for the robot to track the child's progress on certain social and emotional metrics, and report that progress to the child's parents and therapists. In addition, through the companion app, the child's parents are provided with helpful tips in order to continue the learning the child is doing with the robot in the home and apply it to social interactions that occur outside of the home. Different aspects of the STAR Framework, including its evidence-based frameworks, supportive strategies, and metrics are outlined in more detail below.

\subsection{Evidence Based Therapeutic Strategies}

\paragraph{Naturalistic Applied Behavior Analysis.} Applied Behavior Analysis (ABA) is an evidence-based therapy used widely for children with developmental disabilities that focuses on eliciting behavior change or the acquisition of a new skill by providing a series of rewards and consequences, while using strategies that help scaffold and support learning in a way that is tailored to each individual child. Naturalistic ABA (nABA) therapy takes learning new skills out of isolation, and instead incorporates learning into everyday activities the child takes part in, such as reading a book, talking about a favorite animal, or completing a drawing activity. The STAR Framework incorporates several aspects of nABA Therapy, such as having a child work on a specific developmental goal, using scaffolding strategies such as graded cueing to help the child increase success and minimize frustration, providing rewards when a child makes progress towards a goal, and tracking a child's progress towards a goal using metrics and data.

\paragraph{Goal-oriented Activities.} Following the structure of nABA therapy, children will sometimes work on these goals in isolated activities, but will also encounter questions and activities designed to reinforce these goals throughout all activities with the robot. Goals are also broken down into levels to allow for scaffolded learning.

\paragraph{Graded Cueing.} Graded cueing is a strategy where a child is provided with increasingly specific prompts to assist them with answering a question. In the STAR Framework, we have created a system of two different levels of prompts to help a child answer a question. If a child answers a question incorrectly, they are provided with a prompt that allows them to have more support. This system increases success for a child and minimizes frustration.

\paragraph{Cognitive Behavior Therapy.} Cognitive Behavior Therapy, or CBT, is an evidence-based therapy that uses specific verbal strategies to improve emotional regulation and social skills in children with developmental disabilities. While we cannot incorporate all tenets of CBT using a robot medium, we are incorporating evidence-based programs that are designed and based on the principles of CBT into the robot to help children work on emotional regulation. For example, Moxie can teach specific strategies that children can use in order to help calm their body and return to a more natural state. Examples of strategies we can teach children with the robot are Animal Breaths, different mindfulness exercises, and physical movement activities such as dancing.

\paragraph{Adaptive Training.} To serve a diverse population of individuals with varying needs and abilities, Embodied's robotic device requires the ability to continuously track and adapt to the user's abilities and skills. Therefore, our robot leverages the collected multimodal behavior information and responses to standardized, controlled, or unconstrained activities, prompts, or questions. The collected information is compared to expected responses as well as typical behavior collected by other users to assess the individual's abilities and behavior relative to the rest of the observed population. The conceived algorithm leverages a learning algorithm to strategize what activity, prompt, or question to invoke at any given time to maximize the knowledge gained on the user's current abilities as well as to maximize training outcomes by personalizing the selected training \citep{janssen2011motivating,leyzberg2014personalizing,schodde2017adaptive}. The algorithm maintains memory of the scheduled and past activities, prompts, or questions presented to the human user to reduce repetition and accommodate individual preferences. The algorithm maintains knowledge of the human user's abilities and skills to develop a personalized interaction curriculum and to refine an internal model of the user's performance over one or more interactions. The algorithm further compares user performance improvements and how they are related to different activities to select an optimal training plan for each individual user.

\subsection{Content and Engagement Strategy}

It is well known that engagement and adherence to a program are significant drivers of outcomes for digitally administered evidence-informed programs \citep{chin2016successful,michaelides2016weight}. Hence, we designed Moxie as a fun, safe, and engaging companion that helps children build social, emotional, and cognitive skills through everyday play-based learning and captivating content.

Importantly, Moxie is not an entertainment device, toy or on-demand ``information-bot'' but has been purposely designed as a non-gendered peer to children with a curious personality, defined point-of-view, and clear purpose. Namely, Moxie is on a mission from the Global Robotics Laboratory (GRL) to understand how to become a good friend to humans. By giving Moxie personal goals and a defined agenda, we are able to align subject matter and therapy-informed content with the developmental needs of a child to create a personalized engagement strategy where the child takes on the role of being a mentor to the robot. This approach promotes learning through teaching while encouraging important social skills such as developing empathy for others and strategies for successful collaboration.

We approach our content program from a philosophy that Moxie is actively sharing a life journey with a child.  Every week, a different theme such as kindness, friendship, empathy or respect tasks children with missions to help Moxie explore human experiences, ideas, and life skills. These missions are activities that include creative unstructured play like drawing, mindfulness practice through breathing exercises and meditation, reading with Moxie, and exploring ways to be kind to others. Moxie encourages curiosity so children discover the world and people around them so the relationship between Moxie and the child is meant to evolve to allow for changing perspectives, gaining new insights, and modeling a core value of lifelong learning.

Many of Moxie's activities are also designed to help children learn and safely practice essential life skills such as turn taking, eye contact, active listening, emotion regulation, empathy, relationship management, and problem solving.

In terms of experience formatting, we have divided our content experience into the following broad formats of engagement:

\begin{enumerate}
    \item \textbf{One on One Play-with-a-Purpose Activities} designed to help a child develop a specific skill via expert-designed content that provides quantifiable development metrics when a child is playing with the robot.
    \item \textbf{Social Engagement} when the robot facilitates play in order to provide meaningful conversation, social games, personality quizzes or other group activities.
    \item \textbf{Supporting Multimedia Experiences} The ability for a child to participate in associated Moxie-related activities and games via an online G.R.L. kids' portal and monthly physical mailing packs designed to recognize a child's involvement and achievements in the context of a larger creative universe.
\end{enumerate}

Moxie's more specific types of content fit into five broad areas of focus:

\begin{itemize}
    \item \textbf{Stories.}  Reading to/with the robot, creating stories together, and exploring character-based decisions in the context of narrative therapy.
    \item \textbf{Mindfulness.}  Breathing, meditation, mindfulness journeys and other activities to support emotion regulation.
    \item \textbf{Creative Play.}  Imaginative play, drawing, and other creativity-focused content and activities.
    \item \textbf{Life Skills.} Socialization and understanding of important life events (like going to the doctor or a first day at school) and exercises for developing healthy daily habits for living life.
    \item \textbf{GRL Moxie Missions.} Content that encourages a child to explore the world beyond the robot and learn how collaboration, empathy and diversity are keys to creating a better world together.
    \item \textbf{Delighters.}  Unexpected content to provide engagement, such as jokes, riddles or fun facts which creates shareable social currency.
\end{itemize}

Over time, Moxie will also learn more about the child to better personalize its content to help with each child's individual developmental goals and be updated with new content to reinforce the core creative conceit that Moxie is a lifelike companion with its own developmental growth and continuing evolution.

\subsection{Behavioral Metrics}
Traditionally, assessment scales such as the Vineland are administered by a therapist asking the parents or caregivers about the performance or abilities of their child \citep{sparrow2016vineland}. This means that the assessments are subjective, after the fact, and possibly biased. Advances in computational behavior analyses and automatic screening technologies provide an opportunity to profoundly impact the science and treatment of MBDD. Automatic algorithms to detect impairments in social-emotional behavior and functioning from individuals' nonverbal and verbal behavior have the potential to increase quantitative assessments, accessibility, and cost-effectiveness of social-emotional behavior assessment and monitoring. Embodied is uniquely positioned to leverage these new technologies to accurately and quantitatively assess a child's status and progress across many dimensions and over extended periods of time in ecologically valid contexts (i.e., at home in the child's room), during constrained and goal driven interactions, and social play with the robot. This allows us to individualize training for each child and assess progress in transfer tasks.

\begin{table}
  \caption{Tracked Behavioral Metrics}
  \label{tab:behavioral_metrics}
  \centering
  \begin{tabular}{p{3cm}lp{6cm}}
    \toprule
    \textbf{Measurement} & \textbf{Input}  & \textbf{Description} \\
    \midrule
    \rowcolor{LightGray}
    \parbox{3cm}{Engagement} & Multimodal & \parbox{6cm}{Ratio child is engaged and is actively interacting with robot [0, 1]} \\
    \parbox{3cm}{Eye Contact} & Video & \parbox{6cm}{Ratio of time spent looking at interlocutor directly when interacting [0, 1]} \\
    \rowcolor{LightGray}
    \parbox{3cm}{Turn Balance} & Verbal & \parbox{6cm}{Ratio of time spent taking turns during the interaction [0, 1]} \\
    \parbox{3cm}{Conversational Smiles} & Video & \parbox{6cm}{Ratio of time spent smiling while engaging in a conversation [0, 1]} \\
    \rowcolor{LightGray}
    \parbox{3cm}{Social Speech} & Verbal & \parbox{6cm}{Ratio of vocabulary pertaining to friends, family, etc.} \\
    \parbox{3cm}{Relational Speech} & Verbal & \parbox{6cm}{Ratio of use of first person singular pronouns (i.e., my, me, myself, I) vs. plural (i.e., us, we, ours)} \\
    \rowcolor{LightGray}
    \parbox{3cm}{Positive vs Negative Sentiment} & Verbal & \parbox{6cm}{Ratio of positive vs. negative sentiment vocabulary} \\
    \bottomrule
  \end{tabular}
\end{table}

\subsection{Data Storage and Confidentiality}
Embodied obtained Institutional Review Board (IRB) approval for the here presented study and is taking the utmost steps to keep data confidential and secure.

Data was collected by Embodied staff researchers and handled in compliance to Children's Online Privacy Protection Act (COPPA) at all times. Data was recorded in the form of video, audio, transcribed text files, as well as open observation notes and interaction surveys developed by the investigators. The data was encrypted and securely transferred to third party organizations, including an encrypted cloud storage location as well as other cloud-based services (e.g., automatic speech recognition). Only third-party organizations compliant with COPPA were engaged. For encryption, we used 256-bit Advanced Encryption Standard (AES-256) to ensure that the data are protected in transit and at rest. After collection, access to data was restricted to trained and human-subjects certified study personnel only. All participants' data was anonymized and only identified by numbers, not by name. Only CITI-certified personnel has access to numerical codes identifying participant identities.

\section{Experiment and Results}
\label{sec:study}

\subsection{Study Design}
\label{sub:study_design}

\paragraph{Recruitment.}
A single-subject design with replications was conducted to evaluate the effects of repeated interactions with the socially assistive robot Moxie on the social and emotional skills of school-age children with autism spectrum disorders (ASD) or related developmental disorders. All participants were enrolled at the Help Group school in Los Angeles. In total 12 participants (\textit{M} = 9.5, \textit{SD} = 1.6; 11 male) who demonstrated verbally fluent speech (as defined by the capacity to produce multi-word phrases) were recruited.

At baseline the participants were scored on the Social Responsiveness Scale \citet{bruni2014test}, with an average SRS T-score of \textit{M} = 68.6 (\textit{SD} = 6.7), which resulted in an even split of four moderately, four severely, and four mildly impacted participants.

Although we sought to recruit both male and female participants at an equal ratio, we enrolled a greater number of male participants due to the 4:1 ratio of males to females diagnosed with ASD. All families signed informed consent forms.

\paragraph{Study protocol.}
Upon entry into the study, all participants completed a baseline assessment that included standardized parent questionnaires, a brief video-recorded interaction between the child and the robot, a brief video-recorded conversational interaction between the child and an Embodied clinical team member from the study to assess prior knowledge of therapeutic content and social communication skills, and a brief child questionnaire assessing the child's interest in and expectation for a socially assistive robot. Standardized parent questionnaires included the Caregiver Report form of the Vineland Adaptive Behavior Scales \citep{sparrow2016vineland}, Third Edition (VABS-3), the Social Responsiveness Scale, Second Edition \citep{bruni2014test}, and the Social Skills Improvement System Rating Scales \citep{elliot2008social}. Parents also completed two brief questionnaires: one assessing their interest in and expectations for a social robot, and the other comprised simple demographic information. Following the baseline assessment, children  continued with treatment as usual (no experimental intervention will be implemented) for six weeks. Following the six-week baseline period, participants completed an entry assessment. This assessment included all measures from the baseline assessment. In lieu of the complete Caregiver Report form of the VABS-3, parents completed a modified condensed version of the VABS-3 at this time point. All entry assessments were completed within a week of the end of the six-week baseline period. Within a week of the entry assessment, children began the socially assistive robot intervention. Participants engaged with the robot in 15-minute sessions three times per week for six weeks. At the end of the six week intervention, children completed an exit assessment, that included the same measures as the entry assessment. Following the exit assessment, children returned to treatment as usual (no experimental intervention will be implemented) for a period of six weeks. At the end of the six week follow-up period, all participants completed a follow-up assessment that included the same measures as the baseline assessment. A staff member from The Help Group was present with the child at all times throughout the study.

Sessions involved playing games, reading stories, drawing, and talking about specific social skills and emotion regulation strategies. Two of the weekly sessions include 1-on-1 interactions between the child and the robot. The third weekly session includes a peer interaction. Peer interactions involve groups of 2 or 3 children from the same robot interaction group engaging with the robot together. Peers are selected based on coinciding schedules and teacher recommendations; peer groups remain consistent throughout the study. All content for sessions was reviewed by both a licensed occupational therapist and a licensed speech and language pathologist. Sessions were run by one of the licensed therapists. Children were walked to and from their classroom by a study team member or a study team member calls The Help Group staff to walk children to and from sessions, as appropriate. At the end of each session, the participating child completed a brief written survey regarding his or her experience with the socially assistive robot. The session facilitator also completed a brief survey to capture observations of the child's behavior and interest.

\subsection{Results}
\label{sub:results}

In the following we discuss the findings of our study discussed in Section \ref{sub:study_design}. We divide our analysis in subjective assessments following the gold standard questionnaires, such as SRS and SSIS, as well as augment these assessments with quantitative behavior assessments by our tracking and analysis methods.

\begin{figure}
  \centering
  \includegraphics[width=10cm]{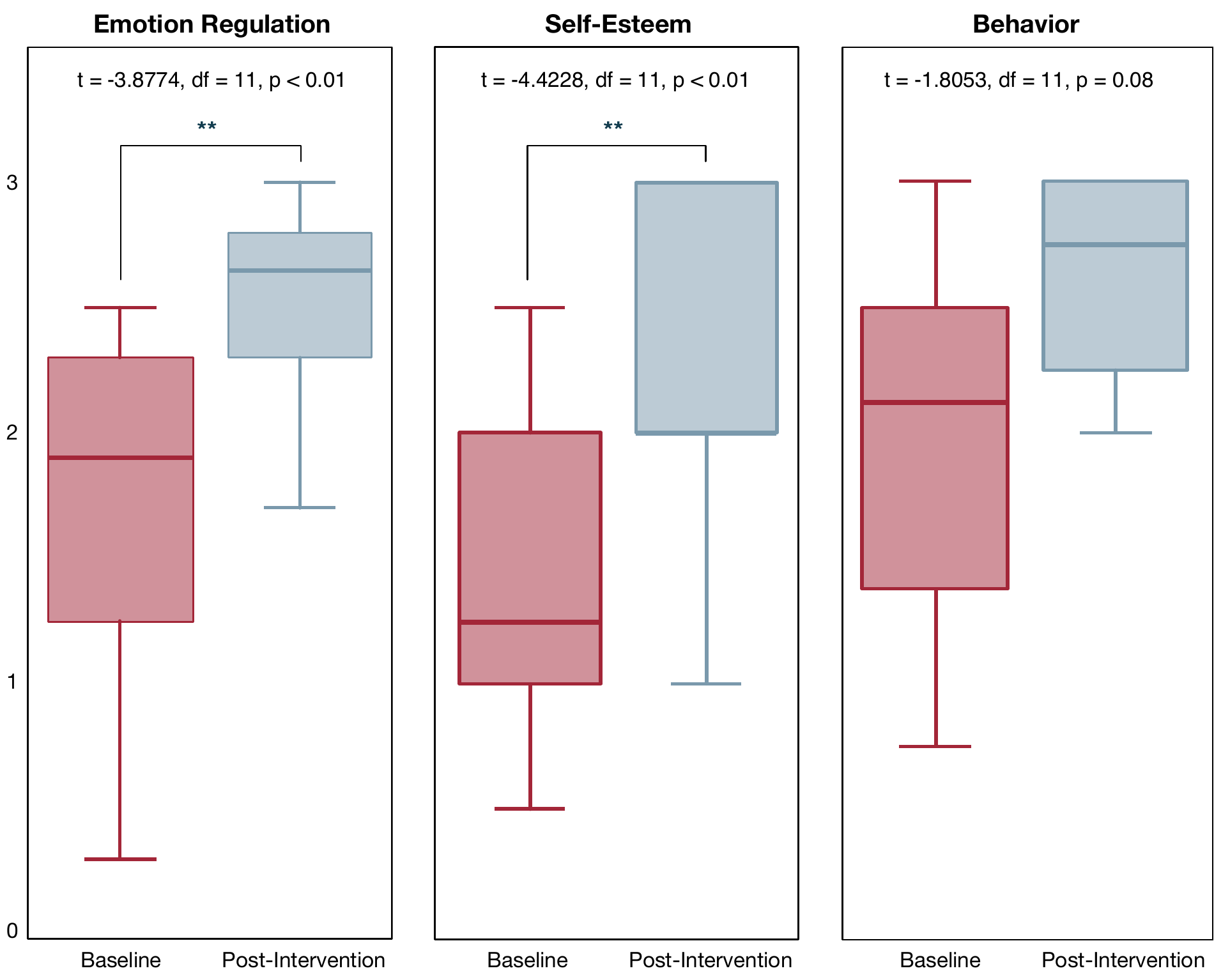}
  \caption{Boxplot of study outcomes for Emotion Regulation, Self-Esteem, and Behavior variables respectively.}
  \label{fig:HG_results1}
\end{figure}

\begin{figure}
  \centering
  \includegraphics[width=10cm]{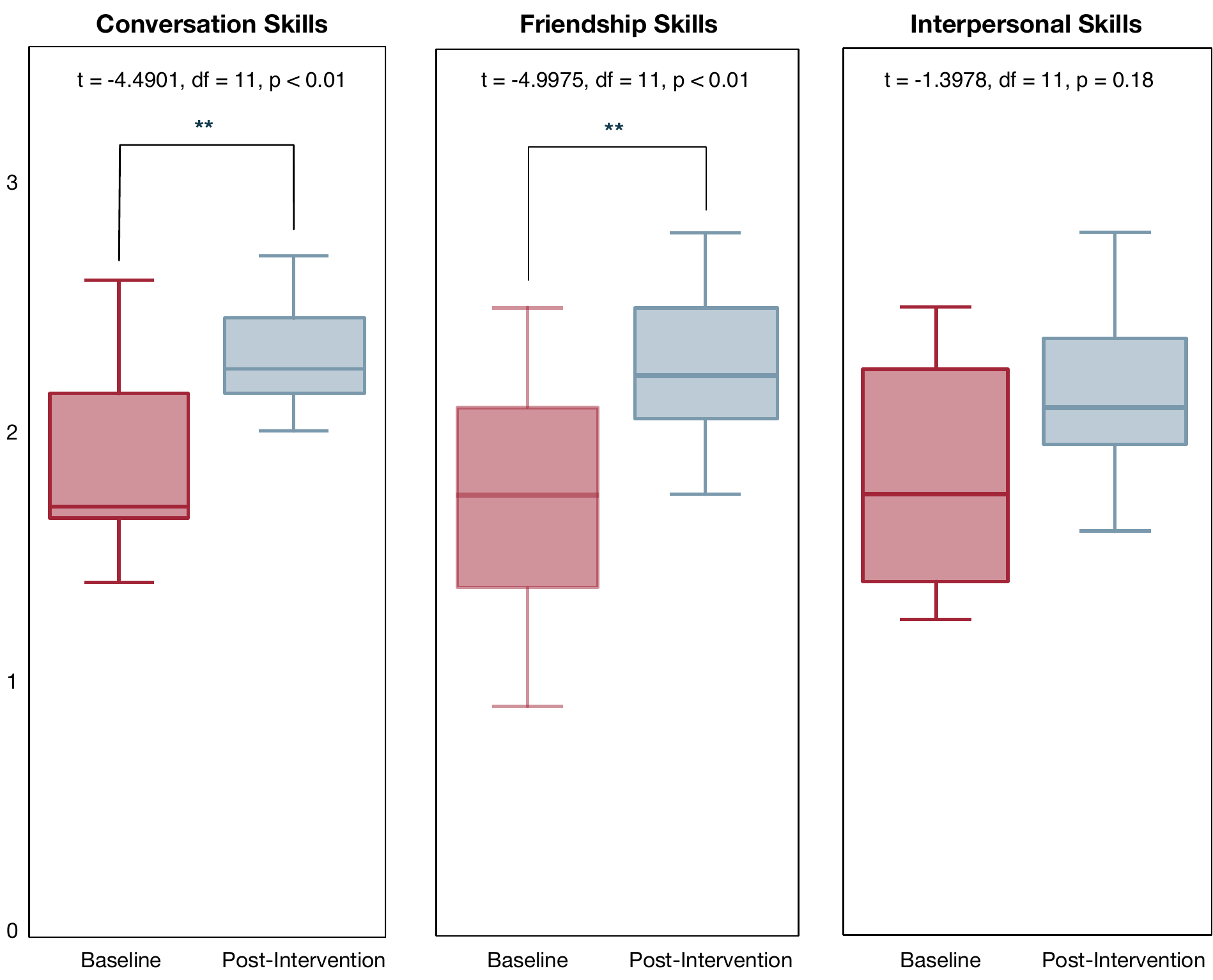}
  \caption{Boxplot of study outcomes for Emotion Regulation, Self-Esteem, and Behavior variables respectively.}
  \label{fig:HG_results2}
\end{figure}

\paragraph{Subjective Assessment.} We collected and grouped six subjective assessments comprised by metrics reflecting scores on the SRS and SSIS scales. The detailed groupings are listed in Table \ref{tab:groupings}. We conducted two-tailed paired t-tests between scores collected at baseline and after the six week intervention. All scores range between $[$0, 3$]$ and are averaged across the grouped items. For \textit{Emotion Regulation} we observe a significant difference between the score at baseline (\textit{M} = 1.75, \textit{SD} = 0.69) and post-intervention (\textit{M} = 2.51, \textit{SD} = 0.39) with t(11) = -3.8774 and p $<$ 0.01. For the group reflecting \textit{Self-Esteem} web also observe a significant difference between the score at baseline (\textit{M} = 1.42, \textit{SD} = 0.60) and post-intervention (\textit{M} = 2.21, \textit{SD} = 0.72) with t(11) = -4.4428 and p $<$ 0.01. For the \textit{Behavior} category we did not observe a significant difference between the score at baseline (\textit{M} = 1.98, \textit{SD} = 0.42) and post-intervention (\textit{M} = 2.48, \textit{SD} = 0.86) with t(11) = -1.8053 and p $=$ 0.08. For \textit{Conversation Skills} we observe a significant difference between the score at baseline (\textit{M} = 1.88, \textit{SD} = 0.37) and post-intervention (\textit{M} = 2.29, \textit{SD} = 0.22) with t(11) = -4.4901 and p $<$ 0.01. For the group reflecting \textit{Friendship Skills} web also observe a significant difference between the score at baseline (\textit{M} = 1.73, \textit{SD} = 0.47) and post-intervention (\textit{M} = 2.20, \textit{SD} = 0.49) with t(11) = -4.9975 and p $<$ 0.01. For the \textit{Interpersonal Skills} category we did not observe a significant difference between the score at baseline (\textit{M} = 1.82, \textit{SD} = 0.47) and post-intervention (\textit{M} = 2.03, \textit{SD} = 0.64) with t(11) = -1.3978 and p $=$ 0.18. Figures \ref{fig:HG_results1} and \ref{fig:HG_results2} summarize the results; baseline results are displayed in red and post-intervention scores are displayed in blue.

\begin{figure}
  \centering
  \includegraphics[width=10cm]{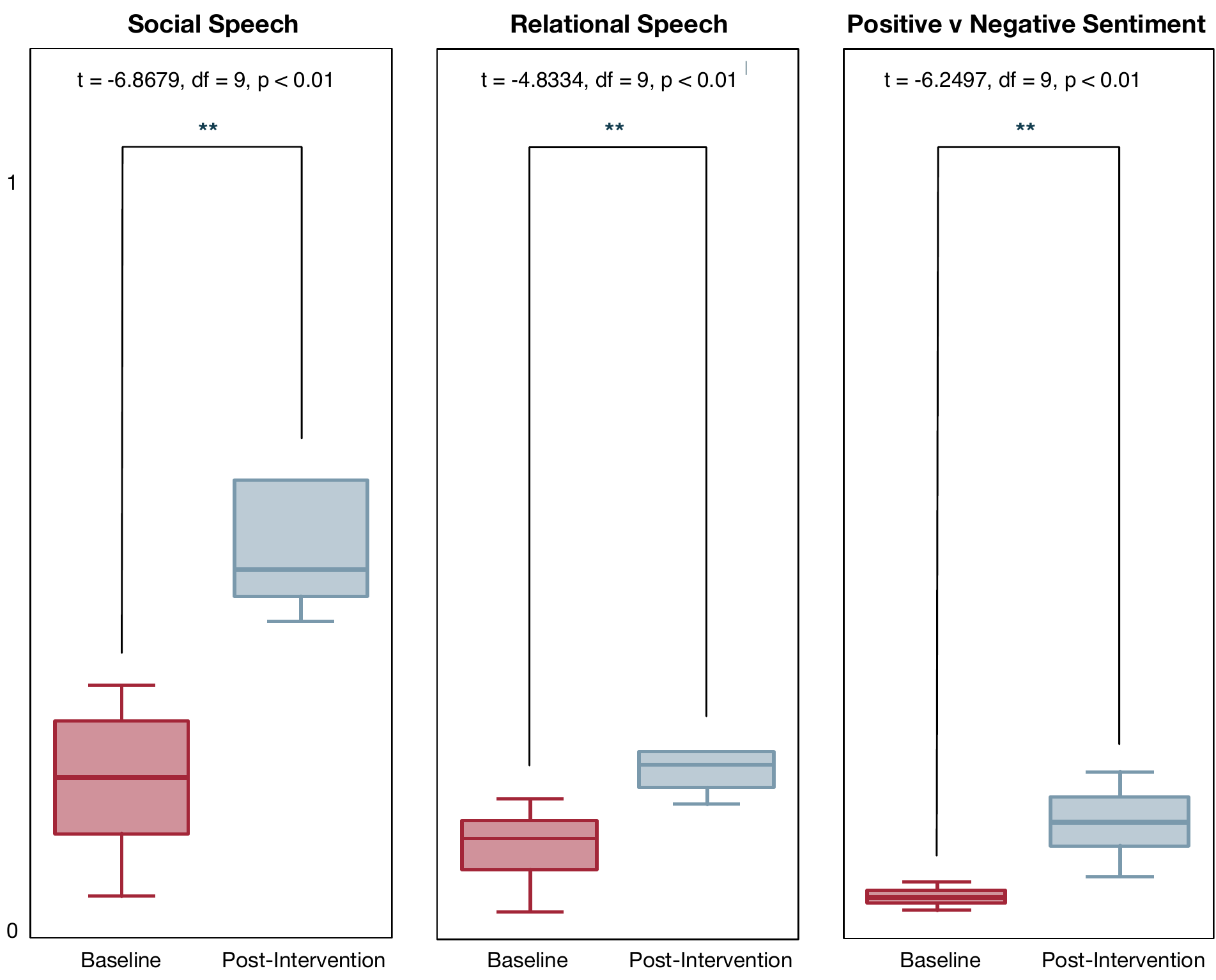}
  \caption{Boxplot of study outcomes for Emotion Regulation, Self-Esteem, and Behavior variables respectively.}
  \label{fig:HG_results4}
\end{figure}

\begin{figure}
  \centering
  \includegraphics[width=10cm]{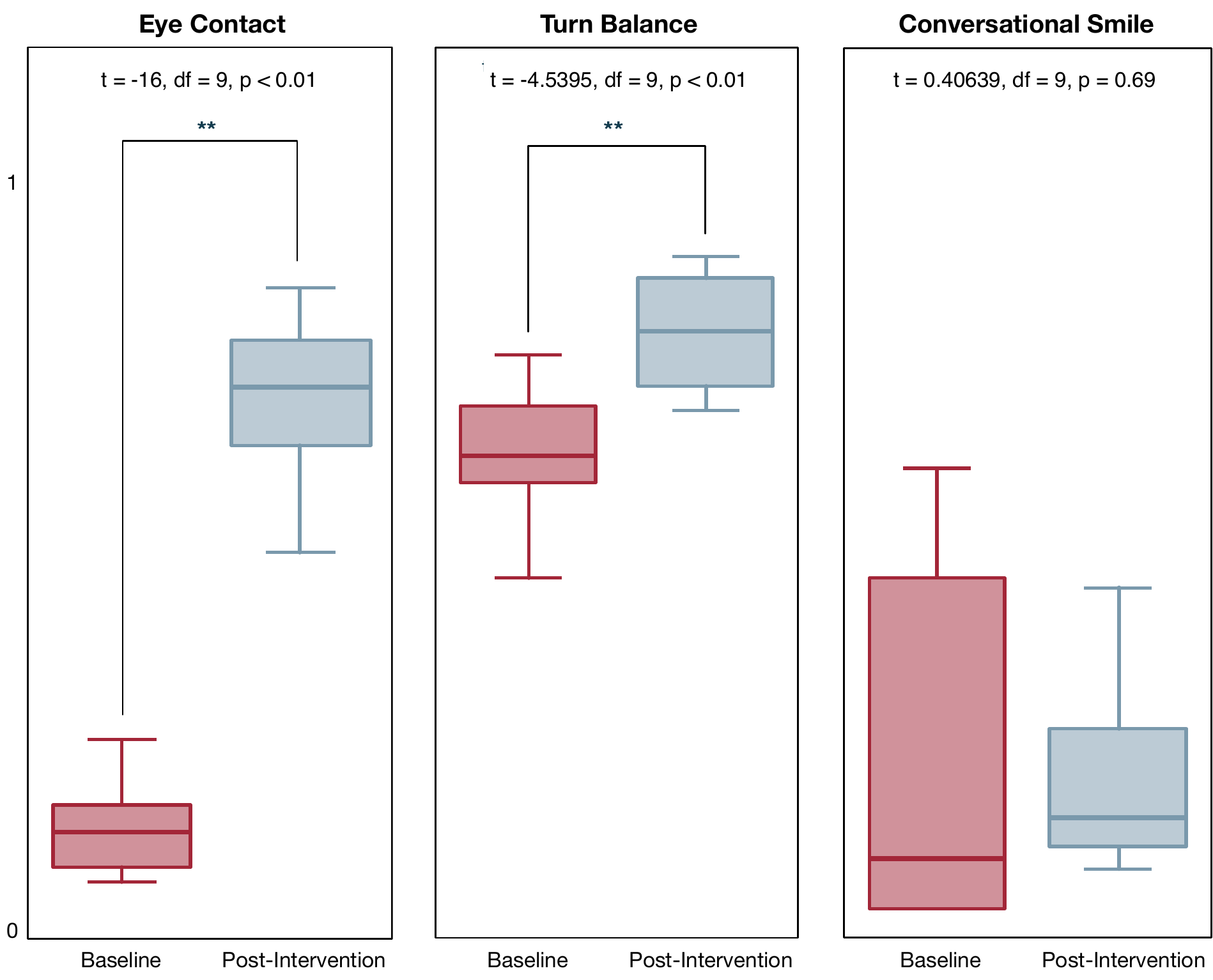}
  \caption{Boxplot of study outcomes for Emotion Regulation, Self-Esteem, and Behavior variables respectively.}
  \label{fig:HG_results3}
\end{figure}

\paragraph{Quantitative Assessment.} We collected six quantitative assessments of the children's behavior and language use, leveraging both facial expression analysis, face tracking, and automatic speech recognition. We conducted two-tailed paired t-tests between the assessed behaviors collected at baseline and after the six week intervention. All scores range between $[$0, 1$]$ and are normalized with respect to the entire interaction for the behavior metrics or all the words spoken within an entire conversation for the language usage metrics. For \textit{Engagement} we observe a significant difference between the score at baseline (\textit{M} = 0.01, \textit{SD} = 0.01) and post-intervention (\textit{M} = 0.52, \textit{SD} = 0.12) with t(9) = -13.27 and p $<$ 0.01. For \textit{Eye Contact} we observe a significant difference between the score at baseline (\textit{M} = 0.14, \textit{SD} = 0.11) and post-intervention (\textit{M} = 0.72, \textit{SD} = 0.11) with t(9) = -16 and p $<$ 0.01. For \textit{Turn Balance} web also observe a significant difference between the score at baseline (\textit{M} = 0.63, \textit{SD} = 0.09) and post-intervention (\textit{M} = 0.81, \textit{SD} = 0.08) with t(9) = -4.5395 and p $<$ 0.01. For the \textit{Conversational Smile} behavior we did not observe a significant difference between the score at baseline (\textit{M} = 0.22, \textit{SD} = 0.26) and post-intervention (\textit{M} = 0.18, \textit{SD} = 0.12) with t(9) = 0.4064 and p $=$ 0.69. For \textit{Social Speech} we observe a significant difference between the score at baseline (\textit{M} = 0.19, \textit{SD} = 0.11) and post-intervention (\textit{M} = 0.57, \textit{SD} = 0.18) with t(9) = -6.8679 and p $<$ 0.01. For the language skill reflecting \textit{Relational Speech} web also observe a significant difference between the score at baseline (\textit{M} = 0.09, \textit{SD} = 0.05) and post-intervention (\textit{M} = 0.24, \textit{SD} = 0.11) with t(9) = -1.8334 and p $<$ 0.01. For the language metric \textit{Positive v Negative Sentiment} we observe a significant difference between the score at baseline (\textit{M} = 0.02, \textit{SD} = 0.02) and post-intervention (\textit{M} = 0.12, \textit{SD} = 0.05) with t(9) = -6.2497 and p $=$ 0.18. Figures \ref{fig:HG_results3} and \ref{fig:HG_results4} summarize the results; baseline results are displayed in red and post-intervention scores are displayed in blue.

\section{Discussion}
Overall, we presented a therapeutic framework, namely the STAR Framework, that leverages established and evidence-based therapeutic strategies delivered by the Embodied Moxie (cf. Figure \ref{fig:moxie}) an animate companion to support children with MBDD. This therapeutic framework jointly with Moxie aims to provide an engaging, safe, and secure environment for children ages 5-10. Moxie delivers content informed by established therapeutic strategies including but not limited to nABA, graded cueing, and CBT. Content ranges from games, drawing, reading, to mindfulness exercises with Moxie. Moxie engages in proactive behavior trying to maximize engagement by the children and leverages an internal curriculum and programming plan to engage each child in activities that maximize their outcome.

While building upon it, the current research surrounding the use of socially assistive robots with children has been largely focused on using robots to provide various types of intervention for the specific needs of children with Autism Spectrum Disorder (ASD). Currently in the literature, there are well over 1,000 peer-reviewed scientific papers that have published consistent evidence for the use of socially assistive robots for children with MBDD. Two recent papers \citep{costa2018comparison,clabaugh2018month} summarize these results concisely:

\begin{itemize}
    \item Socially assistive robots hold a great deal of appeal for children with ASD, and are able to successfully motivate children and sustain their engagement in therapeutic activities that are designed to build social and communicative skills, as confirmed in the here presented work.
    \item Children with ASD who exhibit a wide variety of symptoms and severity levels have all demonstrated sustained positive interactions with socially assistive robots.
    \item Socially assistive robots in these studies varied from looking highly human-like to object-like, however the robots that were perceived by the children to have the most human agency have had the broadest potential for therapeutic effects. 
    \item Improved behavior changes that have been observed in children with ASD following use of a socially assistive robot include increased verbalizations, improved socially appropriate eye gaze, increases in turn taking behaviors, and improved social referencing, perspective taking, sharing, and teamwork.
\end{itemize}

Leveraging multimodal input from a camera and microphones, Moxie assesses a child's progress and struggles throughout the interactions with the robot and reports feedback back to the parents via a companion app (cf. Figure \ref{fig:app}). Moxie is a witness to both skills reflected on standard assessment scales, such as the SRS and SSIS, as well as quantitatively assessed behaviors, such as mutual eye gaze, number of smiles, and language skills.  

In this work, while preliminary, we present our findings of a six-week intervention (see Section \ref{sec:study}). Our results show similar improvements to the list above and findings in the literature \citep{costa2018comparison,clabaugh2018month}. Specifically, we see significant changes in the children's abilities of emotion regulation, self-esteem, conversation skills, as well as friendship skills. These categories span across a broad range of skills and can apply to a number of MBDDs, including ASD, ADHD, and Anxiety, as measured with existing assessment scales. We further complement these subjective assessment scales with metrics directly derived from the children's behavior as it is perceived by the robot. The implemented multimodal behavior analytics system of the Embodied Moxie is in the unique position to directly measure a child's progress and abilities as a first hand witness. We could in fact confirm that children's engagement, eye contact, and turn-balance (i.e., contribution to the interaction) improved significantly as well as their use of social speech and relational speech. Lastly, we also observed an overall more positive sentiment after the intervention.

\section{Conclusions}
Leveraging our STAR framework and the Embodied Moxie within a six week intervention, we found significant improvements both in subjectively assessed skill categories (i.e., emotion regulation, self-esteem, conversation skills, and friendship skills) as well as quantitatively assessed behaviors such as increased engagement, eye contact, contribution to the interaction, social and relational language.

Our findings confirm prior research findings (e.g., \citet{clabaugh2018month,scassellati2018improving}) and are very promising. Due to findings like ours and the overall strength of the evidence in this field, major medical journals have begun publishing results on the use of socially assistive robots for children with ASD, generating a stronger interest in the field from the medical and healthcare communities. In spite of the great promise of these interventions and the large body of previously published work, however, there have been very few studies conducted in the homes of children with ASD, and few that have been measured over an extended period of time. Our future research hopes to expand on previous research in the field by exploring the use of socially assistive robots with children with a range of abilities (including children who may be on the autism spectrum, have ADHD or anxiety, or be neuro-typically developing), and incorporating therapeutic content designed by licensed child development therapists that aims to address a wider variety of social and emotional skills. Specifically, for future work, Embodied seeks to expand its research efforts with an expansive beta test gathering insights in ecologically valid environments (i.e., a child's home rather than in school or at the therapist) over an extended period of time (i.e., 2-3 months intervention).

While Moxie's current content and abilities are already showing promising results, they will grow over time and Moxie's efficacy will improve through findings of future studies. We look forward to improving countless children's abilities through fun and engaging content with Moxie.

\subsubsection*{Acknowledgments}

We would like to thank the robot mentors and their families for their patience and passion for testing Moxie as well as our sponsors on believing in us at Embodied to make this dream come true. For additional information on Moxie and Embodied please visit: \url{www.embodied.com}

\bibliographystyle{apalike}
\bibliography{bibliography}

\clearpage

\appendix 
\section{Appendix A}

\begin{table}[h!]
 \small
  \caption{Groupings for subjective skill assessments.}
  \label{tab:behavioral_metrics}
  \centering
  \begin{tabular}{p{3cm}lp{8cm}}
    \toprule
    \textbf{Category} & \textbf{Questionnaire}  & \textbf{Item} \\
    \midrule
    \rowcolor{LightGray}
    \parbox{3cm}{\textbf{Emotion Regulation}} & SRS & \parbox{8cm}{Seems more fidgety in social situations than when alone.} \\
    \rowcolor{LightGray}
     & SRS & \parbox{8cm}{Has repetitive/odd behaviors such as hand flapping or rocking} \\
    \rowcolor{LightGray}
     & SSIS & \parbox{8cm}{Fidgets or moves around too much.} \\
    \rowcolor{LightGray}
     & SSIS & \parbox{8cm}{Is inattentive.} \\
    \rowcolor{LightGray}
     & SSIS & \parbox{8cm}{Acts anxious with others.} \\
    \rowcolor{LightGray}
     & Vineland & \parbox{8cm}{Changes easily from one activity to another.} \\
    \parbox{3cm}{\textbf{Self-Esteem}} & SRS & \parbox{8cm}{Seems self-confident when interacting with others.} \\
    & SRS & \parbox{8cm}{Has good self-confidence.} \\
    \rowcolor{LightGray}
    \parbox{3cm}{\textbf{Conversation Skills}} & SRS & \parbox{8cm}{Is able to communicate his or her feelings to others.} \\
    \rowcolor{LightGray}
    & SRS & \parbox{8cm}{Gets frustrated trying to get ideas across in conversations.} \\
    \rowcolor{LightGray}
    & SRS & \parbox{8cm}{Thinks or talks about the same thing over and over.} \\
    \rowcolor{LightGray}
    & SRS & \parbox{8cm}{Has trouble keeping up with the flow of a normal conversation.} \\
    \rowcolor{LightGray}
    & SRS & \parbox{8cm}{Has difficulty answering questions directly and ends up talking
around the subject.} \\
    \rowcolor{LightGray}
    & SSIS & \parbox{8cm}{Takes turns in conversations.} \\
    \rowcolor{LightGray}
    & SSIS & \parbox{8cm}{Speaks in appropriate tone of voice.} \\
    \rowcolor{LightGray}
    & SSIS & \parbox{8cm}{Starts conversations with peers.} \\
    \rowcolor{LightGray}
    & SSIS & \parbox{8cm}{Uses gestures or body appropriately with others.} \\
    \rowcolor{LightGray}
    & SSIS & \parbox{8cm}{Repeats the same thing over and over.} \\
    \rowcolor{LightGray}
    & Vineland & \parbox{8cm}{Answers questions that use \textit{what}, \textit{where}, \textit{who}, and \textit{why}.} \\
    \rowcolor{LightGray}
    & Vineland & \parbox{8cm}{Says what he/she knows or thinks about things.} \\
    \rowcolor{LightGray}
    & Vineland & \parbox{8cm}{Talks with the right loudness, speed, and excitement for the convo.} \\
    \rowcolor{LightGray}
    & Vineland & \parbox{8cm}{Talks with others about shared interests.} \\
    \rowcolor{LightGray}
    & Vineland & \parbox{8cm}{Stays on topic in conversation when needed.} \\
    \rowcolor{LightGray}
    & Vineland & \parbox{8cm}{Talks with others about things they are interested in.} \\
    \parbox{3cm}{\textbf{Friendship Skills}} & SRS & \parbox{8cm}{Avoids eye contact or has unusual eye contact.} \\
    & SRS & \parbox{8cm}{Has difficulty making friends, even why trying his/her best.} \\
    & SRS & \parbox{8cm}{Avoids starting social interactions with peers or adults.} \\
    & SRS & \parbox{8cm}{Focuses his/her attention to where others are looking/listening.} \\
    & SRS & \parbox{8cm}{Has a sense of humor/understands jokes.} \\
    & SRS & \parbox{8cm}{Is too tense in social settings.} \\
    & SRS & \parbox{8cm}{Stares or gazes off into space.} \\
    & SSIS & \parbox{8cm}{Says \textit{thank you}.} \\
    & SSIS & \parbox{8cm}{Takes care when using other people's things.} \\
    & SSIS & \parbox{8cm}{Responds well when others start a conversation or activity.} \\
    & SSIS & \parbox{8cm}{Makes eye contact when talking.} \\
    & Vineland & \parbox{8cm}{Makes good eye contact when he/she interacts with people.} \\
    & Vineland & \parbox{8cm}{Answers politely.} \\
    & Vineland & \parbox{8cm}{Keeps a proper distance from others in social situations.} \\
    & Vineland & \parbox{8cm}{Is a good friend.} \\
    & Vineland & \parbox{8cm}{Talks with others without interrupting or being rude.} \\
    \rowcolor{LightGray}
    \parbox{3cm}{\textbf{Behavior}} & SSIS & \parbox{8cm}{Follows directions/instructions.} \\
    \rowcolor{LightGray}
    & SSIS & \parbox{8cm}{Disobeys rules or requests.} \\
    \rowcolor{LightGray}
    & SSIS & \parbox{8cm}{Lies or does not tell the truth.} \\
    \rowcolor{LightGray}
    & SSIS & \parbox{8cm}{Cheats in games or activities.} \\
    \parbox{3cm}{\textbf{Interpersonal Skills}} & SRS & \parbox{8cm}{Is aware of what others are thinking or feeling.} \\
    & SRS & \parbox{8cm}{Is able to understand the meaning of other people's tone of voice
and facial expressions.} \\
    & SRS & \parbox{8cm}{Offers comfort to others when they are sad.} \\
    & SRS & \parbox{8cm}{Responds appropriately to mood changes in others.} \\
    & SSIS & \parbox{8cm}{Tries to make others feel better.} \\
    & SSIS & \parbox{8cm}{Tries to comfort others.} \\
    & Vineland & \parbox{8cm}{Understands the meaning of at least 3 facial expressions.} \\
    & Vineland & \parbox{8cm}{Realizes when others are happy, sad, angry, etc.} \\
    \bottomrule
  \end{tabular}
\end{table}

\end{document}